\documentclass[letterpaper, 10 pt, conference]{ieeeconf}
\IEEEoverridecommandlockouts		
\usepackage{balance}

\usepackage{amsmath}
\usepackage{amssymb}
\usepackage{pifont}

\usepackage{tabularx}
\usepackage{graphicx}
\usepackage{bm}
\usepackage{color}
\usepackage{float}
\usepackage{units}
\usepackage{comment}
\usepackage{hyperref}

\makeatletter 
\def\endfigure{\end@float} 
\def\endtable{\end@float}
\makeatother 

\usepackage{subfig}
\usepackage{algorithm}
\usepackage{algorithmic}

\renewcommand{\unit}[1]{{\rm #1} }

\newcommand{\edit}[1]{\textcolor{black}{#1}}

\renewcommand{\vec}[1]{ {\mathbf{#1}}}

\usepackage{multirow}
{\end{tabbing} \end{mdframed} \vspace{5px}}

\newcommand{\BM}{\begin{bmatrix}}
\newcommand{\EM}{\end{bmatrix}}

\newcommand{\beq}{\begin{equation}}
\newcommand{\eeq}{\end{equation} }

\IEEEoverridecommandlockouts

\begin{document} 

\title{\Large \bf 			
Continuous Jumping for Legged Robots on Stepping Stones via Trajectory Optimization and Model Predictive Control
	\\[-1ex]
}
	
\author{Chuong Nguyen$^{*}$, Lingfan Bao$^{*}$, and Quan Nguyen 
\thanks{The authors are with the Department of Aerospace and Mechanical Engineering, University of Southern California, USA: {\tt\small vanchuong.nguyen@usc.edu, lingfanb@usc.edu, quann@usc.edu }}%
\thanks{$^*$ Contributed equally to this work}%
}%

\maketitle

\begin{abstract}
Performing highly agile dynamic motions, such as jumping or running on uneven stepping stones has remained a challenging problem in legged robot locomotion. This paper presents a framework that combines trajectory optimization and model predictive control to perform robust and consecutive jumping on stepping stones. In our approach, we first utilize trajectory optimization based on full-nonlinear dynamics of the robot to generate periodic jumping trajectories for various jumping distances. A jumping controller based on a model predictive control is then designed for realizing smooth jumping transitions, enabling the robot to achieve continuous jumps on stepping stones. Thanks to the incorporation of MPC as a real-time feedback controller, the proposed framework is also validated to be robust to uneven platforms with unknown height perturbations and model uncertainty on the robot dynamics. 
\end{abstract}

\section{Introduction}
\label{sec:Introduction}
The capability of navigating uneven terrain with discrete footholds such as stepping stones or stairs is a remarkable advantage of legged robots over their wheeled counterparts. This advantage has recently attracted much attention, resulting in considerable studies on walking legged robots on stepping stones using either control
 (\cite{Quan_bipedal_17},\cite{Quan_bipedal_3Dterrain_16},\cite{Quan_bipedal_terrain_18}) or learning (\cite{Deepgait_quadruped},\cite{2020-ALLSTEPS},\cite{Avinash_perception_walking},\cite{Takahiro_learningloco_2022}) frameworks. \edit{ However, the realization of highly dynamic locomotion on stepping stones, such as jumping, has not been well explored}. 



Related works on jumping for legged robots \cite{matthew_mit2021_1},\cite{HWPark2021},\cite{YaranDing_2020_planning_MICP} including our prior works \cite{QuannICRA19},\cite{chuong_jumping_3D_2021},\cite{guillaume_jump_2020} only consider a single jump.
\edit{Our work \cite{QuannICRA19} introduces a trajectory
optimization framework based on full-body dynamics to allow MIT Cheetah 3 to perform a single jump
on a high platform. However, no feedback control on the body motion is considered to compensate for errors of the jumping trajectory. Therefore, the framework is not robust to jumping from an uneven platform. 
Another offline trajectory optimization approach is proposed in \cite{Winkler_RAL2018} to find feasible motions over discrete terrain in simulation. Nevertheless, this method relies on simplified dynamics which does not consider leg dynamics and robot actuator constraints. In addition, there is also no feedback control embedded in this work.
These simplifications may limit the accuracy as well as the success rate of transferring dynamic jumping motions to the robot hardware. 
Recently, combinations of a single jump with multiple bounding are also designed to enable a robot to jump over obstacles on flat terrain in \cite{HWPark2021}, and jump over gaps along discrete terrain in \cite{Gabriel_learning_jump_2021}.
Departing from prior works, we are interested in consecutive jumping on stepping stones. Our framework utilizes the combination of nonlinear trajectory optimization and model predictive control to realize robust and continuous jumping of legged robots on stepping stones}.


Unlike walking (\cite{Quan_bipedal_17}-\cite{Avinash_perception_walking}), the jumping motions on stepping stones come with additional challenges including (1) high efficient jumping transitions, (2) hard impact on the environment, and (3) long aerial time. In particular, jumping transitions occur in a very short contact time and on a limited terrain surface. The hard impact requires efficient real-time feedback controllers to control the ground reaction force (GRF) to mitigate its perturbation on the whole body motion. 
In addition, because the robot motion has little impact on the body trajectory during a flight phase, a small error in the body motion during the jumping phase could result in a significant error in the landing phase as well as in the subsequent jumps. These problems thus make continuous jumping highly challenging.

\begin{figure}[!t]
	\centering
	{\centering
		\resizebox{\linewidth}{!}{\includegraphics[trim={0cm 1cm 0cm 0cm},clip]{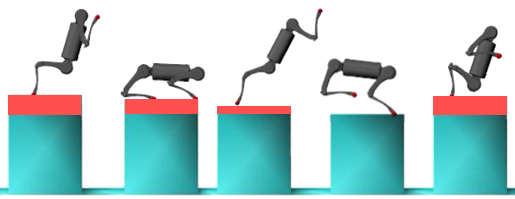}}
	}
	\caption{
	Continuous jumping of quadruped robots over stepping stones with random distance and unknown height perturbation. Supplemental video: \url{https://youtu.be/jBGY1K1UbhM}.}
	\label{fig:double_roll}
\end{figure}

\begin{figure*}[!t]
	\centering
	\resizebox{0.9\linewidth}{!}{\includegraphics[trim={0cm 5cm 0cm 4cm},clip]{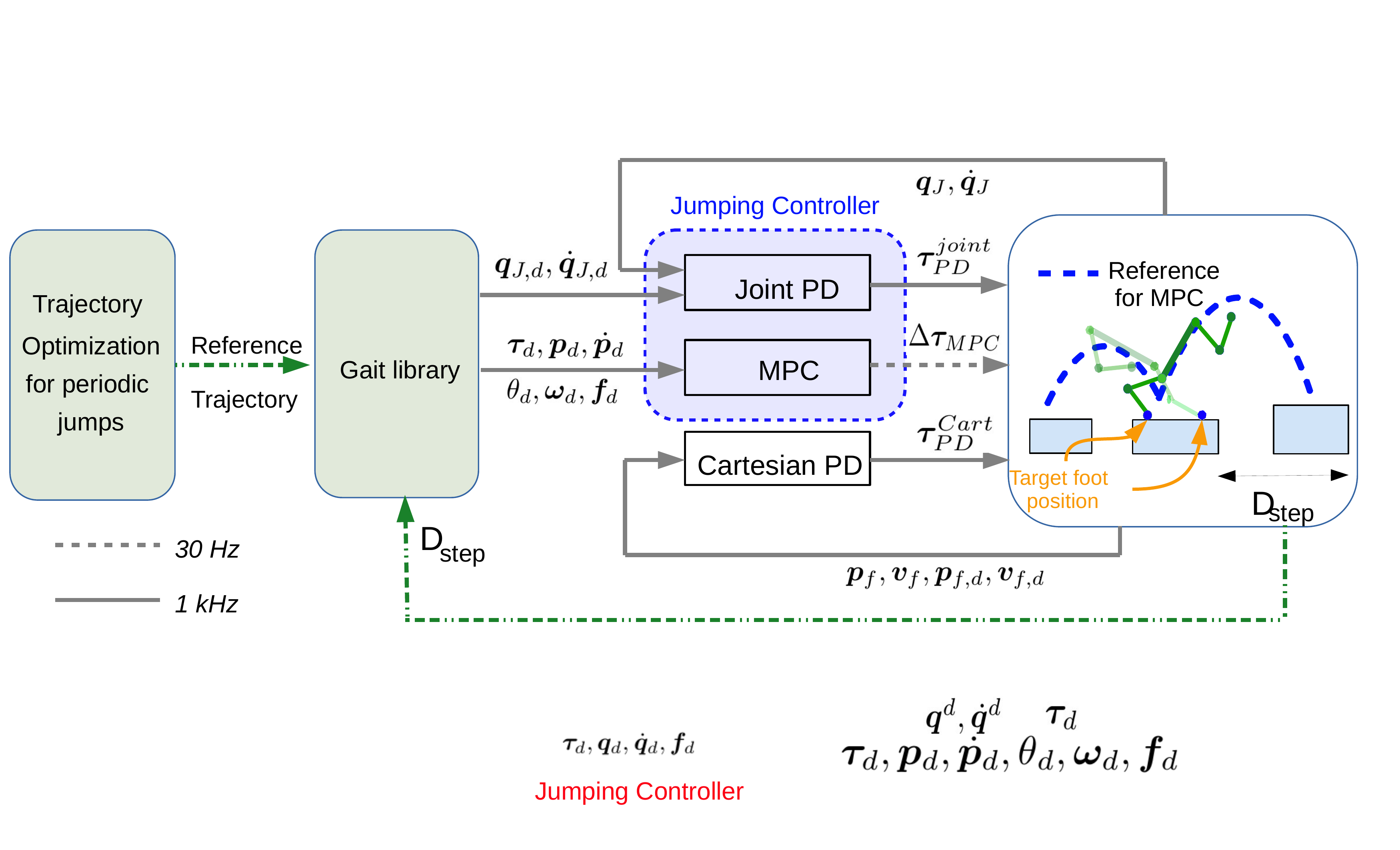}}
	\caption{Block diagram for the proposed framework.} 
	\label{fig:framework}
\end{figure*}





Continuous jumping on stepping stones requires whole-body coordination while respecting all constraints of robots' actuators. Therefore, designing an
approach to optimize for the whole-body motion is crucial. The optimization is also essential to maximize the jumping performance
to enable the robot to traverse significant gaps.
Thus, we design a trajectory optimization (TO) framework to generate optimal periodic gait references, which will then be used for continuous jumps. The optimization framework adopts the full-body dynamics of the robot to leverage the whole-body motion for jumping while satisfying all physical constraints.


Model Predictive Control (MPC) has recently been widely
used in legged robots’ locomotion thanks to its capability to realize robust locomotion over a wide variety of gaits or contact modes
(\cite{Carlo2018},\cite{Donghyun19},\cite{GerardoBledt}).
However, in these works, the MPC formulation is designed for stabilizing gaits with short flight time and based on the assumption of small variations of the body orientation. Alternatively, we design a jumping controller based on MPC to achieve continuous jumping transitions and to tackle a wide range of body orientation and long flight phases.
In this paper, we combine MPC and joint PD controller to track the reference motion from the trajectory optimization module.
This combination allows
the robot to jump consecutively on challenging
stepping stones at high accuracy. In addition, it shows the robustness to unknown height perturbation of the platform and model uncertainty (e.g., carrying an unknown load).


The contribution of our work is summarized as follows:

\begin{itemize}
    \item We propose a framework that allows quadruped robots to perform online consecutive jumping on stepping stones. A trajectory optimization based on full-body dynamics is formulated to generate a set of references for periodic jumping. The gait library is then designed to achieve an online gait generation for different jumping distances. 
    \item A MPC-based jumping controller is designed to efficiently handle jumping transitions between different jumps and guarantee accurate jumps
on stepping stones.
    
    
    
    
    \item Our framework is validated on the A1 robot model jumping on a variety of terrains: randomly-placed stepping stones, unknown perturbation of terrain height, unknown load carrying, and their combinations.
    \item Our experiments in hardware validate the robustness of the jumping controller \edit{for a single jump} on uneven terrain
    with unknown height perturbation.
\end{itemize}

The rest of the paper is organized as follow.
An overview of the proposed framework is presented in Section \ref{sec:success_jump_framework}. The trajectory optimization framework and a gait library are described in Section \ref{sec:optimization}.  Section \ref{sec:JumpingControl} presents a proposed controller for jumping. Results from hardware experiments and simulation are shown in Section~\ref{sec:Results}. Finally, Section~\ref{sec:Conclusion} provides concluding remarks.





\section{Overview of the framework}
\label{sec:success_jump_framework}

In this section, we introduce the overview of our proposed
approach. A block diagram of our framework is illustrated
in Figure. \ref{fig:framework}. 
Firstly, trajectory optimization is used to generate reference trajectories for different periodic jumps. It is then combined with a gait library to generate online references for jumping of different distances. Secondly, a feedback controller based on MPC and joint PD is designed to robustly track the reference model and to realize smooth transitions between different jumps. 

The trajectory optimization is formulated and solved off-line for a small number of gait references. The gait interpolation policy \cite{Quan_bipedal_terrain_18} computes a new reference for the next jump prior to landing on the next stone based on the actual measured distance of $D_{step}$. 
This reference is updated to the jumping controller: $\{\bm{\tau}_d, \bm{p}_d, \dot{\bm{p}}_d, \theta_d , \bm{\omega}_d, \bm{f}_d\}$ for MPC, and $\{\bm{q}_{J,d}, \dot{\bm{q}}_{J,d}\}$ for joint PD controller. 
The MPC and joint PD controller are updated at $30$ $Hz$ and $1$ $kHz$ respectively. 

During a transition to a next jump, it is important to enforce precise footstep placement on stepping stones in order to keep the robot on the terrain and avoid accumulated errors. Therefore, we utilize a Cartesian PD controller to drive each swing foot to the predefined targets on stepping stones. The controller executes $\bm{\tau}_{PD}^{Cart}$ at $1~kHz$.

\section{Trajectory Optimization and Gait Library}
\label{sec:optimization}

\subsection{Periodic Jumping Gait Optimization}
Due to the limitation of the robot's actuators, it's critical to leverage the whole body motion to maximize the jumping capability of the robot.
Moreover, due to periodic patterns in continuous jumps, the motions can be formulated as connections of periodic jumping gaits. 
In this section, we propose an optimization framework to generate a certain number of periodic gaits, which will be extended to consecutive jumps with different jumping distances. We also utilize full-body dynamics to generate high accuracy jumping references and optimal whole-body coordination, while respecting all physical constraints of the robot.


\subsubsection{Dynamical Model for Jumping} For quadrupedal animals, high jumping is normally restricted to a sagittal plane. 
Therefore, this paper will also focus on 2D motion of the robot. The robot model thus can be considered as a rigid-body system consisting of $5$ links in the 2D plane, and the equation of motion is formulated as follows \cite{QuannICRA19}:  

\begin{equation}\label{eq:full_dynamics_2D}
\begin{bmatrix}\bm{M} & -\bm{J}_c^T \\ -\bm{J}_c^T & \mathbf{0} \end{bmatrix} \begin{bmatrix}
\bm{\ddot{q}} \\ \bm{f}_c
\end{bmatrix}= \begin{bmatrix}-\bm{C}\bm{\dot q}-\bm{g} + \bm{S}\bm{\tau}+\bm{S}_{f}\bm{\tau}_{f} \\ \bm{\dot J}_{c}(\bm{q})\bm{\dot q}\end{bmatrix}
\end{equation}

where $\bm{q}:=\left[x;~z;~q_{pitch};~\bm{q}_{J}\right]$ is a vector of generalized coordinates, in which $x,z,q_{pitch}$ are the CoM position and body's pitch angles respectively, and $\bm{q}_J$ is a vector of joint angles. $\bm{M}$ is the mass matrix, $\bm{C}$ is represented for Coriolis  and centrifugal terms, $\bm{g}$ denotes gravity vector, $\bm{J}_c$ is the spatial Jacobian expressed at the contact foot, $\bm{S}$ and $\bm{S}_{fric}$ are distribution matrices of actuator torques $\bm{\tau}$ and the joint friction torques $\bm{\tau}_{fric}$, $\bm{f}_c$ is the spatial force at the contact feet. The dimension of $\bm{J}_c$ and $\bm{f}_c$ depend on the phase of gait, and the number of legs in contact with the ground.
In addition, we also denote positions of front feet as $\bm{p}_{F}=[p_F^x;p_F^z]$, and positions of rear feet as $\bm{p}_R=[p_R^x;p_R^z]$.

\begin{figure}[!t]
	\centering
	{\centering
		\resizebox{0.9\linewidth}{!}{\includegraphics[trim={0cm 5.5cm 0cm 2.8cm},clip]{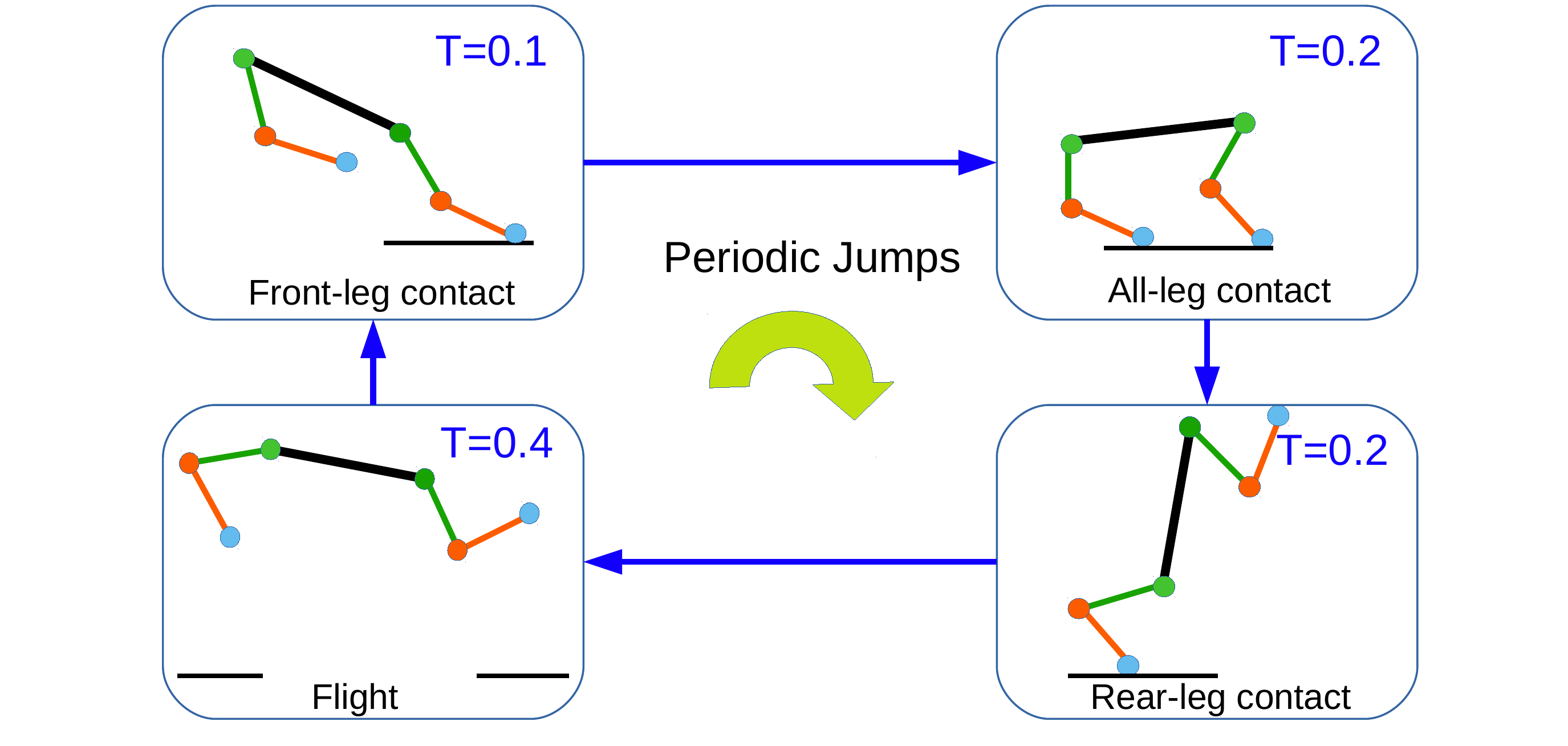}}
	}
	\caption{\textbf{Periodic jumping phases}. Each periodic jumping motion consists of four phases}
	\label{fig:contact_schedule}
\end{figure}

\subsubsection{Contact Schedule}
Inspired by nature, we formulate each jump consisting of four sequential phases: front-leg contact, double contact, rear leg contact, and flight phase. All phases are solved via off-line optimization. To establish a periodic jump, the initial pose of the first phase is identical to the final pose of the last phase (see Fig. \ref{fig:contact_schedule}).

\subsubsection{Cost function and Constraints}
The periodic trajectory optimization is formulated as follows:
\begin{subequations} \label{eq:time_opt}
\begin{align}
\textrm{min}& J=\sum_{k=1}^{N}w_q \|\bm{q}_{J,k}-\bm{q}_{J,ref}\|_2^2+w_{\tau} \|\bm{\tau}_k\|_2^2 \nonumber\\
\text{s.t. }
             & ~ ~ ~ ~ ~ ~ ~ ~ ~ ~ ~ ~ ~ ~ ~  \textrm{ } \textrm{\textbf{Periodic constraints}:} \nonumber \\
             & x_{N}=x_{1}+D_{des}, ~ z_{N}=z_{1}, ~q_{pitch,N}=q_{pitch,1}, \nonumber \\
             & ~ ~ ~ ~ ~ ~ ~ ~ ~ ~ ~ ~   \bm{q}_{J,N} = \bm{q}_{J,1},~ \dot{\bm{q}}_{N}=\dot{\bm{q}}_{1} \label{eq:time_opt_1} \\
             & ~ ~ ~ ~ ~ ~ ~ ~ ~ ~ ~ ~ ~ ~ ~ ~ \textrm{ } \textrm{\textbf{Foot position}:} \nonumber \\
             & ~ ~\bm{p}_{F,k}=\bm{0}, ~\textrm{in front-leg and all-leg contact}  \label{eq:time_opt_2}  \\
             & \bm{p}_{R,k}=[-d;0], ~ \textrm{in all-leg and rear-leg contact}  \label{eq:time_opt_3}  \\
             &  ~ ~ ~ ~ ~ ~ ~ ~ ~ ~ ~ ~ ~ ~ ~ ~ p_{F,k}^z \geq 0, p_{R,k}^z \geq 0 \label{eq:time_opt_4}\\
             & ~ ~ ~ ~ ~ ~ ~ ~ ~  \textrm{\textbf{Full-body dynamics constraints} (\ref{eq:full_dynamics_2D})}\\
             & ~ ~ ~ ~ ~ ~ ~ ~ ~ ~\textrm{ } \textrm{\textbf{Pre-landing configuration}:} \nonumber \\
             & ~ ~ ~ ~ ~ ~ ~ ~ ~ ~   \bm{\dot q}_{J,k}=\bm{0}~ (k\in[(N-10):N])  \label{eq:time_opt_5} \\
             &\textrm{ } \textrm{\textbf{Joint angle limits}:} ~ \bm{q}_{\bm{J},min}\le \bm{q}_{\bm{J},k} \le \bm{q}_{\bm{J},max}  \label{eq:time_opt_6} \\
             & \textrm{ } \textrm{\textbf{Joint angular velocity limits}:} ~ |\bm{\dot q}_{\bm{J},k}|\le \bm{\dot q}_{\bm{J},max}  \label{eq:time_opt_7}  \\
             & ~ ~ ~ ~ ~ ~  ~ ~ ~  ~ ~ ~  \textrm{ } \textrm{\textbf{Torque limits}:} |\bm{\tau}_k|\le \bm{\tau}_{max}  \label{eq:time_opt_8} \\
             &  ~ ~ ~  ~ ~ ~  ~ ~ ~ \textrm{\textbf{GRF limits}:} \bm{f}_{min}^z \le \bm{f}_{k}^z \le \bm{f}_{max}^z  \label{eq:time_opt_9} \\
             & ~ ~ ~ ~ ~ ~ ~ ~ ~ \textrm{ } \textrm{\textbf{Friction cone limits}:} |\bm{F}_k^x/\bm{F}_k^z|\le \mu \\
             & ~ ~ ~ ~ ~ ~ ~ ~ ~ ~ ~ ~ ~ ~ ~  \textrm{\textbf{Geometric  constraints}}
\end{align}
\label{eq:timings_TO_of_SRB}
\end{subequations}

where $\bm{q}_{J,k}$ is a joint angle, and $\bm{\tau}_k$ is a joint torque at the iteration $k^{th}$; $w_q, w_{\tau}$ are corresponding weights of these optimization variables. 
Note that since the primary goal of the optimization is to maximize the performance of the robots to jump over a large gap between stepping stones, we do not over-regulate the use of joint torques. Therefore, we use a dominant weight for the joint positions in comparison with a weight for torque (e.g. $w_{\tau}=0.005$ $w_q=1$).
For foot position constraints in (\ref{eq:time_opt_2})\&(\ref{eq:time_opt_3}), $d$ is pre-defined distance between the front and rear feet.
Geometric constraints are imposed to guarantee: (a) the robot body and legs have a good clearance with terrain, and (b) each robot part does not collide with others.

Having presented an optimization approach to generate individual periodic jumping gait, we will next design a gait library to generate online jumping references.

\subsection{Jumping Gait Library}
\label{sec:gait_interpolation}
 In order to  adapt to the change of terrain structure quickly, it is important to have a policy to update the reference for full-body motion quickly. However, since solving the TO for full-body dynamics in real-time is not applicable due to the complexity of the problem, we design a gait library and gait interpolation policy \cite{Quan_bipedal_17},\cite{Quan_bipedal_terrain_18} to update the reference at the beginning of each jump quickly. 
 
 The optimization framework in the previous section is used to generate a gait library consisting of four periodic jumping gaits with jumping distances of $D_{step}^* = \{0.6, 0.7, 0.8, 0.9\}~m$. 
Having this gait library, we then do gait interpolation to get the desired jumping gait with an arbitrary step length between these discrete values, $D_{step,i}^*, 1 \leq i \leq 4$. 

In particular, for each nominal step length $D_{step,i}^*, 1 \leq i \leq 4$, the reference trajectory $\bm{Q}_i^*=[\bm{q}; \dot{\bm{q}}; \bm{\tau}; \bm{f}]$ is utilized for linear interpolation. The result trajectory $\bm{Q}$ for each random value of step length $D_{step} \in [ D_{step,i}^*,  D_{step,i+1}^*] $, is computed as
\begin{subequations}\label{eq:gait_1}
    \begin{align}
    &\gamma(D_{step})=\frac{D_{step}-D_{step,i}^*}{D_{step,i+1}^*-D_{step,i}^*}, \\
    \bm{Q}(D_{step})&=(1-\gamma(D_{step}))\bm{Q}_i^*+ \gamma(D_{step})\bm{Q}_{i+1}^*.
    \end{align}
\end{subequations}

\section{Jumping Controller}
\label{sec:JumpingControl}
\begin{figure*}[!t]
	\centering
	\subfloat[\textbf{Case I - Random distance}. The stepping stone are randomly placed with step length in the large range from $60 cm$ to $90 cm$]{\centering
		\resizebox{0.9\linewidth}{!}{\includegraphics[trim={0cm 0.1cm 0.2cm 0.3cm},clip]{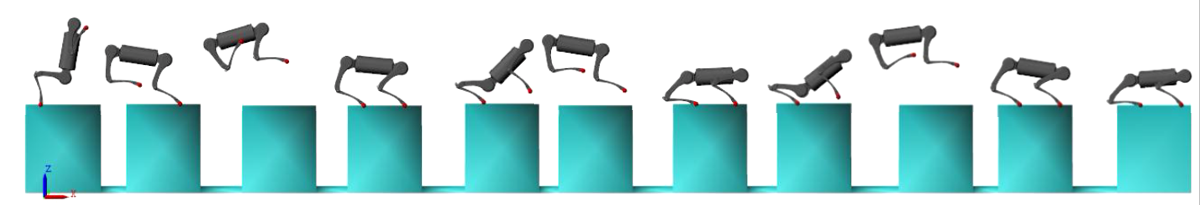}}}\label{fig:terrain_a}\\
	\subfloat[\textbf{Case II - Random distance and random height perturbations}. The stepping stone are randomly placed with step length in $(60:90) cm$. The gap heights vary randomly between $-6 cm$ to $+5 cm$, which are unknown to the robot]{\centering
		\resizebox{0.9\linewidth}{!}{\includegraphics[trim={0.1cm 0.1cm 0cm 0.3cm},clip]{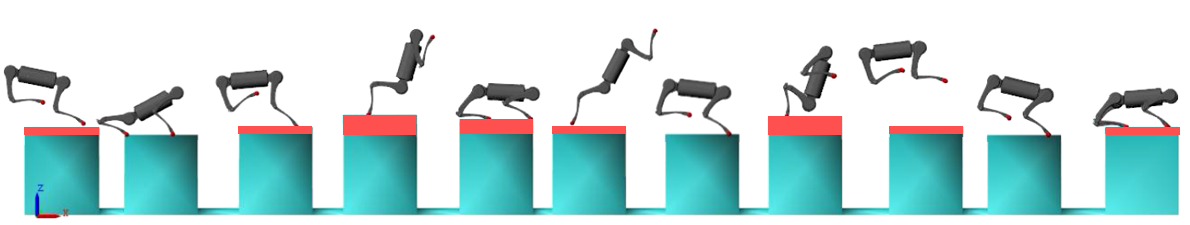}}}\label{fig:terrain_b}\\
	\subfloat[\textbf{Case III - Random distance and unknown mass}. The stepping stones are randomly placed with step length in the range of $60:90(cm)$. The robot carries a load of $2kg$, up to $17\%$ of robot weight, which is unknown to the controller]{\centering
		\resizebox{0.9\linewidth}{!}{\includegraphics[trim={0cm 0.1cm 0cm 0.2cm},clip]{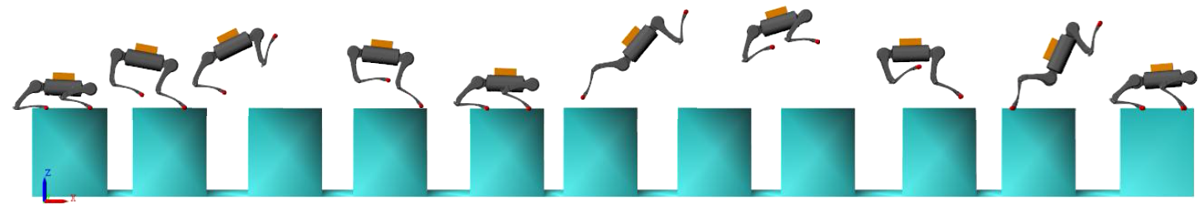}}}\\
	\label{fig:terrain}
	\caption{\textbf{Simulation: } The robot performs continuous jumps on uneven stepping stones. Video: \url{https://www.youtube.com/watch?v=jBGY1K1UbhM}}
	\label{fig:terrain}
\end{figure*}
In this section, we design a jumping controller to realize continuous jumps. Our proposed controller combines MPC and joint PD controller to track the reference trajectory from the optimization. This combination allows the robot to achieve smooth jumping transitions and accurate jumping trajectories on uneven stepping stones.

\subsubsection{Jumping Controller}
Firstly, we revisit a simplified rigid body dynamics model of quadruped robots in the vertical plane as follows:
\begin{subequations}\label{eq:simplified dynamics}
    \begin{align}
    \ddot{\bm{p}} &=\frac{\sum_{i=1}^{2}\mathbf{f}_{i}}{m}-\mathbf{g}, \\
    \frac{\mathrm{d}}{\mathrm{d} t}(I{\bm{\omega}}) &=\sum_{i=1}^{2} \mathbf{r}_{i} \times \mathbf{f}_{i},
    \end{align}
\end{subequations}

where $\bm{p}$ is the CoM position in the world frame; $\bm{r}_i=[r_{ix};r_{iz}]$ and $\bm{f}_i=[f_{ix};f_{iz}]$ denotes the position of contact point relatively to CoM, and contact force of foot $i^{th}$ respectively in the world frame; $\bm{\omega}=\dot{\theta} \vec{k}$ is angular velocity of the body; $\theta$ is a pitch angle. Then by define $\bm{x}=[\bm{p};\theta;\dot{\bm{p}};\dot{\theta},g]$, $g$ is gravity, $\bm{f}=[\bm{f}_1; \bm{f}_2]$, the dynamics is rewritten as:
\begin{equation}\label{eq:simplified robot dynamics}
    \dot{\bm{x}}(t)=\bm{A}_{c} \mathbf{x}(t)+\bm{B}_{c}\left(\mathbf{r}_{1}, \ldots, \mathbf{r}_{n} \right) \bm{f}(t),
\end{equation}
where $\bm{A}_{c}= \begin{bmatrix}
\bm{0}_{3\times3} & \bm{I}_{3\times 3} & \bm{0}_{3\times1}\\
\bm{0}_{3\times3} & \bm{0}_{3\times 3} & \bm{e}_g\\
\bm{0}_{1\times3} & \bm{0}_{1\times 3} & 0
\end{bmatrix}$, $\bm{e}_g= \begin{bmatrix} 0 & -1 & 0
\end{bmatrix}^\top$,

$\bm{B}_{c}=\begin{bmatrix}
\bm{0}_{3\times2} & \bm{0}_{3\times2}\\
\bm{I}_{2 \times 2}/m & \bm{I}_{2 \times 2}/m \\
I^{-1} [\bm{r}_1]_{\times} & I^{-1} [\bm{r}_2]_{\times} \\
\bm{0}_{1\times2} & \bm{0}_{1\times2}
\end{bmatrix}$, $[\bm{r}_i]_{\times}=[-r_{iz}, r_{ix}]$. \\


which will then be formulated in discrete time:
\begin{equation}\label{eq:simplified robot dynamics_discrete}
    \bm{x}_{k+1}=\bm{A}_k \mathbf{x}_{k}+\bm{B}_k \bm{f}_{k}.
\end{equation}

The works in \cite{Carlo2018} and \cite{GerardoBledt} initially propose a MPC to keep the robot balanced while performing locomotion with the short flight time. This paper, on the other hand, designs a controller specifically for jumping to control a wide range of the body orientation and to achieve long flight phases.

Highly agile jumping motions on stepping stones require accurate tracking of the jumping reference with a wide range of body motion.
An error in body position or orientation before taking off usually ends up with a significant deviation from reference upon landing. 
In order to ensure a high tracking performance for the jumping motion in real time, we combine MPC and a joint PD control to follow the reference trajectory generated from Section \ref{sec:optimization}.

In particular, the MPC problem is formulated as a quadratic programming (QP) with moving horizons to solve for optimal GRFs, which minimizes the weighted tracking errors of the body's trajectory reference and the GRF reference obtained from Section \ref{sec:optimization}:
\begin{subequations} \label{eq:MPC_jumping}
\begin{align}
 \textrm{min} \textrm{ } & \sum_{k=t}^{t+N-1} \| \bm{x}_k-\bm{x}_{d,k} \|_{P_k} + \| \bm{f}_k-\bm{f}_{d,k} \|_{Q_k},\\
\text{s.t. }
             & \bm{x}_{k+1}=\bm{A}_k\bm{x}_k+ \bm{B}_k \bm{f}_k, \forall k=t, ..., t+N-1 \label{eq:MPC_jumping_c1} \\
             & \bm{c}_k^{min} \leq \bm{C}_k \bm{x}_k \leq \bm{c}_k^{max}, \forall k=t, ..., t+N-1 \label{eq:MPC_jumping_c2} \\ 
             & \bm{D}_k \bm{x}_k =\bm{0}, \label{eq:MPC_jumping_c3} \forall k=t, ..., t+N-1
\end{align}
\end{subequations}
where a contact force reference $\bm{f}_d$ is obtained from trajectory optimization in Section \ref{sec:optimization}; $P_k$ and $Q_k$ are the weighted diagonal matrices at step $k$; $N$ is a number of predicted horizon. The equation (\ref{eq:MPC_jumping_c2}) captures constraints related to friction cone and force limits, while the equation (\ref{eq:MPC_jumping_c3}) is represented for the shifting contact schedule.
For the reference in moving horizons, when a predicted horizon exceeds the take off time (at the end of rear-leg contact), the reference of states and contact forces at this horizon takes the values at the take-off time. To improve the tracking performance, in the cost function (\ref{eq:MPC_jumping}), we also enforce dominant weights regarding to states and contact force components at the last horizon compared to other horizons. This allows us to put more weights on minimizing the errors before taking off.

The above problem (\ref{eq:MPC_jumping}) can be reformulated as a dense form of constrained QP as follows:
\begin{subequations} \label{eq:MPC_jumping_to_QP}
\begin{align}
 \textrm{min} \textrm{ } & \frac{1}{2} \bm{F}_t^T \bm{H}_t \bm{F}_t + \bm{F}_t^T \bm{b}_t,  \\
\text{s.t. }
             & \bm{c}_t^{min} \leq \bm{C}_t\bm{F}_t \leq \bm{c}_t^{max}, \label{eq:MPC_jumping_to_QP_c1}
\end{align}
\end{subequations}
where 
\begin{subequations}\label{eq:MPC_jumping_to_QP_term}
\begin{align} 
             \bm{H}_t& =2 \bm{B}_{qp,t}^T \bm{S} \bm{B}_{qp,t}+ 2 \bm{\alpha} \label{eq:MPC_jumping_to_QP_c1} \\
             \textrm{ } \textrm{ } \bm{b}_t& = 2 \bm{B}_{qp,t}^T \bm{S} (\bm{A}_{qp,t}\bm{x}_t-\bm{X}_{t,d})- 2 \bm{\alpha}^T \bm{F}_{t,d} \label{MPC_jumping_to_QP_c2}
\end{align}
\end{subequations}
Here, $\bm{A}_{qp,t}$ and $\bm{B}_{qp,t}$ are constructed from $\bm{A}_k$ and $\bm{B}_k$ ($\forall k=t,...,t+N-1$), $\bm{x}_t$ is a current state at time step $t$, and  $\{\bm{c}_t^{min},\bm{c}_t^{max}\}$ represents inequality constraints on the GRF. $\bm{X}_{t,d}$ and $\bm{F}_{t,d}$ are the reference of states and GRF, which concatenate references in the considered moving horizons from $t$ to $t+N-1$. These references are obtained from trajectory optimization in Section \ref{sec:optimization}.
The readers can refer \cite{Juan_2011} for more details on how to formulate the MPC as the constrained QP in general.

The solution $\bm{f}_{MPC}^*$ that takes the value at the first horizon of solution $\bm{F}_t$ in (\ref{eq:MPC_jumping_to_QP}) will be utilized to compensate for the errors between actual and reference jumping:
\begin{align} \label{eq:delta_MPC}
    \Delta \bm{\tau}_{MPC}=\bm{J}(\bm{q}_{j})^\top \bm{R}^\top[\bm{f}_{MPC}^* - \bm{f}_{d}],
\end{align} 
where $\bm{J}(\bm{q}_{j})$ is the foot Jacobian at the configuration $\bm{q}_{j}$; $\bm{R}$ is the rotation matrix which transforms from body
to world frame. The compensation $\Delta \bm{\tau}_{MPC}$ will be combined with the joint PD controller, leveraging the reference of the torque $\bm{\tau}_d$ and joint profiles $ \{\bm{q}_{J,d}, \bm{\dot q}_{J,d} \}$ obtained from the trajectory optimization in Section \ref{sec:optimization}. This results in a feed forward torque applying to the robot actuators:
\begin{align} \label{eq:MPC_feedforward_torque}
    \bm{\tau}_{ff}=\Delta \bm{\tau}_{MPC}+ \bm{\tau}_d + \bm{\tau}_{PD}^{joint},
\end{align} 
\begin{align}
\bm{\tau}_{PD}^{joint} = \bm{K}_{p,j}(\bm{q}_{J,d}-\bm{q}_J)+\bm{K}_{d,J}(\bm{\dot q}_{J,d}-\bm{\dot q}_J)
\end{align} \label{eq:jointPDController}
\edit{This combination allows us to achieve accurate tracking performance and robust to uncertainties, which will be validated in Section \ref{sec:Results}}.
\subsubsection{Jumping Transitions}
Jumping transitions play a crucial role in guaranteeing successful continuous jumps on stepping stones.
To achieve high efficiency jumping transitions, we combine the MPC-based jumping controller with a Cartesian PD controller for foot placement, which is illustrated in Fig. \ref{fig:framework}. 
The jumping controller applies the feed forward torque to the contact legs, and the torque value is computed as (\ref{eq:MPC_feedforward_torque}).
For the swing legs, we utilize the Cartesian PD controller for accurate foot placement on the next stepping stones as follows
\begin{align}
\bm{\tau}_{PD}^{Cart}=\bm{J}(\bm{q}_{j})^\top  \bm{R}^\top[\bm{K}_{p}(\bm{p}_{f,d}-\bm{p}_d) +\bm{K}_{d}(\bm{v}_{f,d} - \bm{v}_f)], \nonumber
\end{align} \label{eq:cartesian_controller}
where $\bm{K}_p$ and $\bm{K}_d$ are diagonal matrices of proportional and derivative gains; $\bm{p}_{f}$ and $\bm{v}_{f,d}$ are actual foot position and velocity measured in the world frame; and the target foot position $\bm{p}_{f,d}$ on the next stepping stone in world frame is set at the beginning of a pre-landing configuration.

We reapply MPC for the stance legs when that legs impact the ground (end of the flight phase). Note that the time when the other leg touches the ground is usually earlier or later than expected, causing a mismatch between the predicted contact schedule and actual contact states. For jumping on stepping stones with a very limited contact time, this issue accumulates errors along with the motions, affecting the accuracy of the next jumping. To improve the accuracy, we enforce the availability of actual contact states in the MPC. In particular, at the time all legs have impact with the ground, we recompute the MPC started with double contact.



\section{Results}
\label{sec:Results}

\begin{table}[t!]
	\centering
	\caption{A1 Robot Parameters}
	\label{tab:PRP}
	\begin{tabular}{cccc}
		\hline
		Parameter & Symbol & Value & Units\\
		\hline
		Max Torque & $\tau_{max}$    & 33.5 & $\unit{Nm}$  \\[.5ex]
		Max Joint Speed & $\dot{q}_{max}$    & 21 & $\unit{rad}/\unit{s}$  \\[.5ex]
		Total robot mass & $m$    & 12 & $\unit{kg}$  \\[.5ex]
		Trunk dimension & $l,w,h$ & 0.361, 0.194, 0.114 & $\unit{m}$ \\[.5ex]
		Trunk Inertia  & $I_{xx},I_{yy},I_{zz}$  & 0.017, 0.056, 0.065 & $\unit{kg}.\unit{m}^2$ \\[.5ex]
		Leg Link Lengths & $l_{1}, l_{2}$ & 0.2 & $\unit{m}$ \\[.5ex]
		\hline 
	\end{tabular}
\end{table}

\subsection{Numerical Simulation}
We validate the effectiveness of our framework on the A1 robot model with parameters and its actuation capabilities summarized in Table \ref{tab:PRP}.
We use the open-source optimization toolbox Casadi to set up and solve the trajectory optimization for periodic jumps, then design simulation platforms based on Matlab-Simscape. Three different simulation cases are considered for jumping on stepping stones as follows:

\begin{itemize}
    \item \textit{Case I- Random distance }(Fig.\ref{fig:terrain}a): The distance between two adjacent stepping stones is selected as:
\begin{equation}
L_{d}={[67,78,71,79,63,77,70,86,67,80]}(\mathrm{cm}) \nonumber
\end{equation}
    \item \textit{Case II- Random distance and random height perturbations} (Fig.\ref{fig:terrain}b): 
\begin{align}
&L_{d}= {[67,78,71,79,63,77,70,86,67,80]}(\mathrm{cm}), \nonumber \\
& {h_{d} = [-4,0,5,3,0,-6,4,0,-5,0](\mathrm{cm})}. \nonumber
\end{align}
The difference of the height of the stones $h_d$ is measured relatively to the original height of the first stone. Note that the height perturbation is unknown to the robot. We also use the same distances between stepping stones as in  the Case I to validate the robustness of our framework on the unknown structure of the terrain.
   \item \textit{Case III- Carrying unknown load} (Fig.\ref{fig:terrain}c). The robot carries a load of $2kg$, about $17\%$ of the robot's weight, which is unknown to the controllers. The stepping stone distances are also set up similarly to the Case I to show the efficiency of our controllers in compensating for the unknown disturbance introduced to the robot model.
\end{itemize}

\begin{figure}[!t]
	\centering
	\subfloat[CoM trajectory]{\centering
		\resizebox{0.9\linewidth}{!}{\includegraphics[trim={0cm 0cm 0cm 0cm},clip]{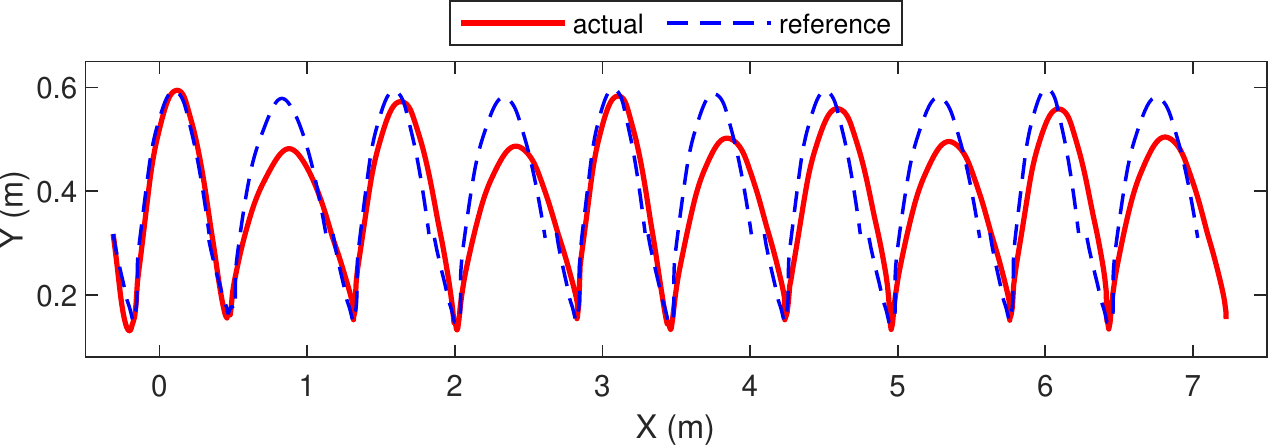}}}\\
	\subfloat[Pitch angle]{\centering
		\resizebox{0.9\linewidth}{!}{\includegraphics[trim={0cm 0cm 0cm 0cm},clip]{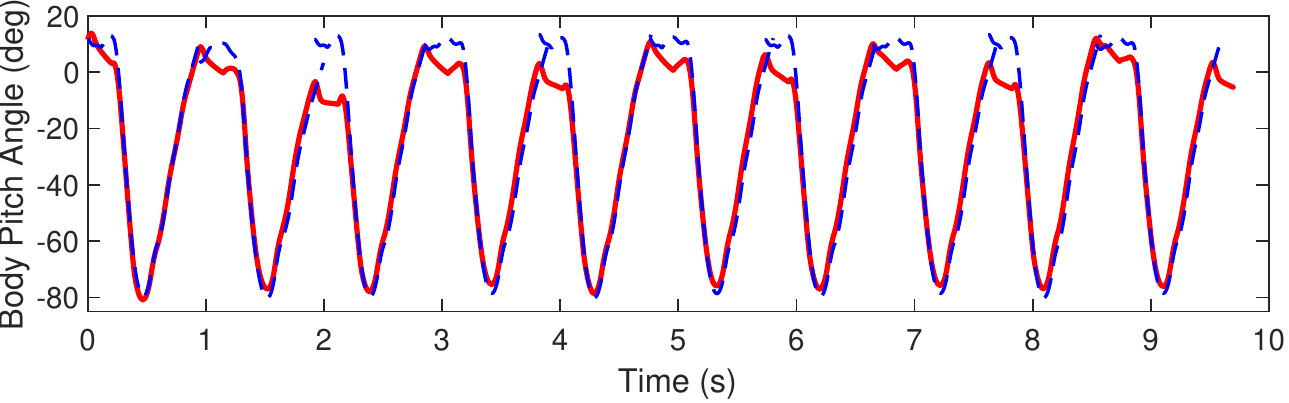}}}\\
	\caption{\textbf{Case I- Tracking performance of the jumping controller} The A1 robot successively jumps on stepping stones with a stochastic gap distance.} 
	\label{fig:jumping_random_length}
\end{figure}

Fig. \ref{fig:jumping_random_length} shows the tracking performance of our design controllers for Case I. Our controllers ensure the high accuracy tracking of body orientation and position during contact phases, as well as enable the robot to traverse over stepping stones at high accuracy. We believe that our work is the first that successfully demonstrates continuous jumping on stepping stones for the quadruped robots.

\begin{figure}[!t]
	\centering
	\resizebox{0.9\linewidth}{!}{\includegraphics[trim={0cm 0cm 0cm 0cm},clip]{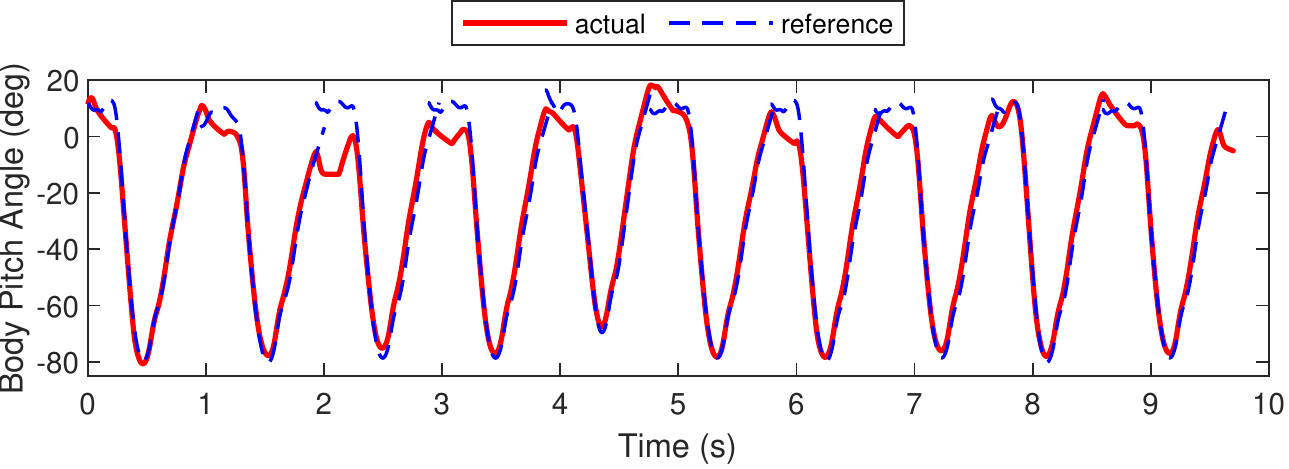}}
	\caption{\textbf{Case II- Tracking performance of the jumping controller.} The robot jumps on the stepping stones with random distance and unknown random height perturbations.}
	\label{fig:Jumping_random_distance_random_height}
\end{figure}

\begin{figure}[!t]
	\centering
	\subfloat[Front leg]{\centering
		\resizebox{\linewidth}{!}{\includegraphics[trim={0cm 0cm 0cm 0cm},clip]{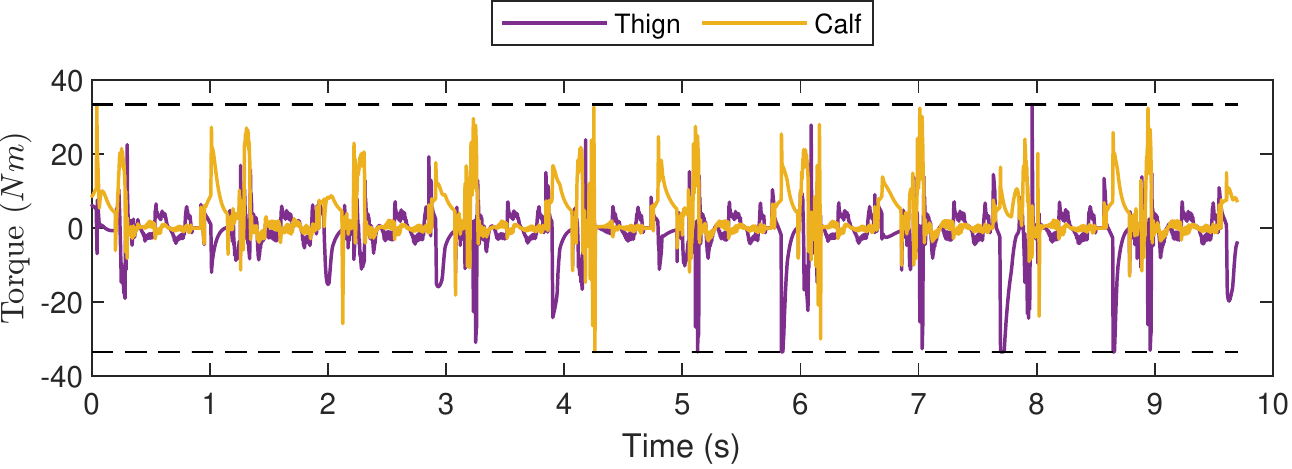}}}\\
	\subfloat[Rear leg]{\centering
		\resizebox{\linewidth}{!}{\includegraphics[trim={0cm 0cm 0cm 0cm},clip]{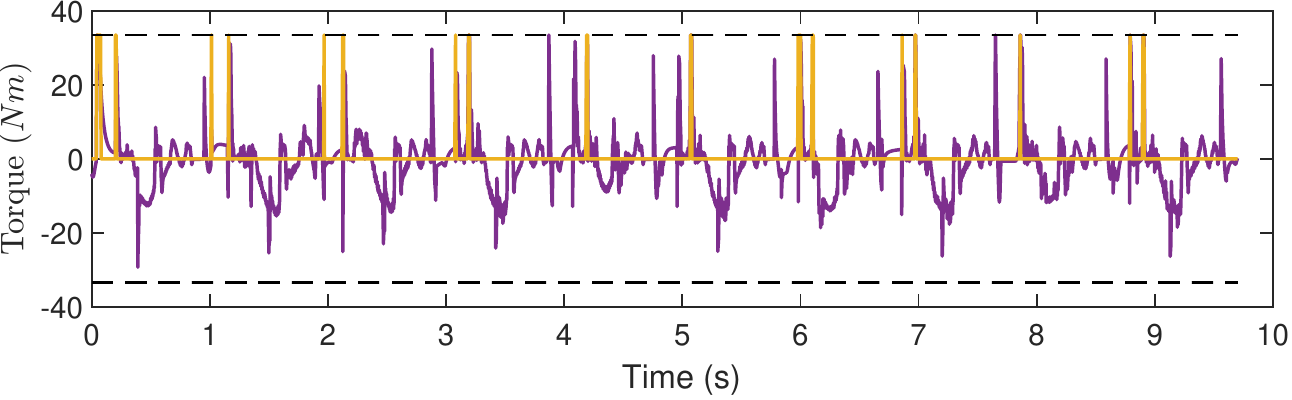}}}\\
	\caption{\textbf{Case II- Torque profile.} The actuators' torques are within the limits during consecutively jumping motions.} 
	\label{fig:controller_force_random_length_random_height}
\end{figure}

\begin{figure}[!t]
	\centering
	\resizebox{0.9\linewidth}{!}{\includegraphics[trim={0cm 0cm 0cm 0cm},clip]{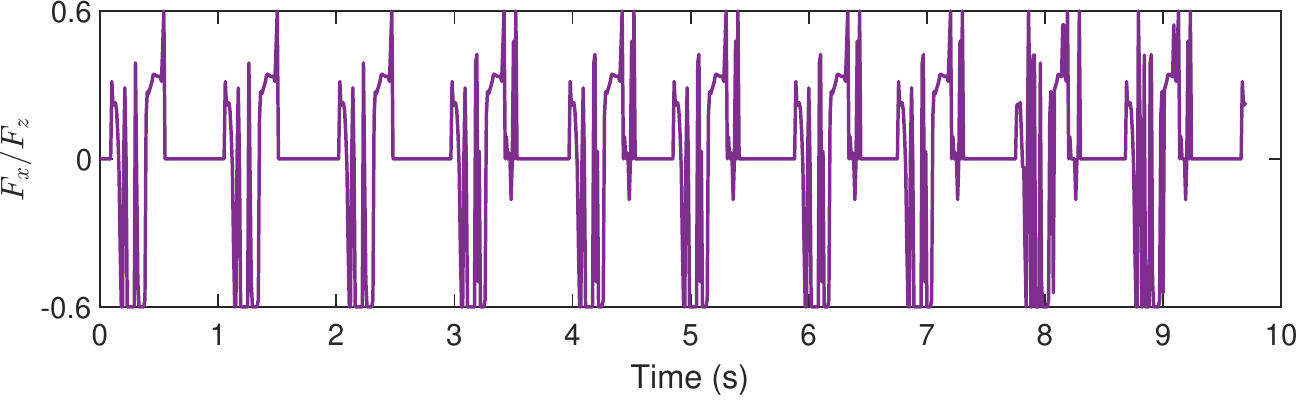}}
	\caption{\textbf{Case II- Friction cone limits.} The force commands satisfies $|F_x/F_z|\leq0.6$ to prevent slippery during jumping motions.}
	\label{fig:controller_friction_force_random_length_random_height}
\end{figure}

For Case II, Fig.\ref{fig:Jumping_random_distance_random_height} shows that even with unknown fluctuation in the height of stepping stone, our controller is still able to track the orientation reference and guarantee successful jumps. Fig.\ref{fig:controller_force_random_length_random_height} and Fig.\ref{fig:controller_friction_force_random_length_random_height} show that the torques satisfy the actuation limits, and the outputs of the MPC are within friction cone limits. 

\begin{figure}[!t]
	\centering
	\resizebox{0.9\linewidth}{!}{\includegraphics[trim={0cm 0cm 0cm 0cm},clip]{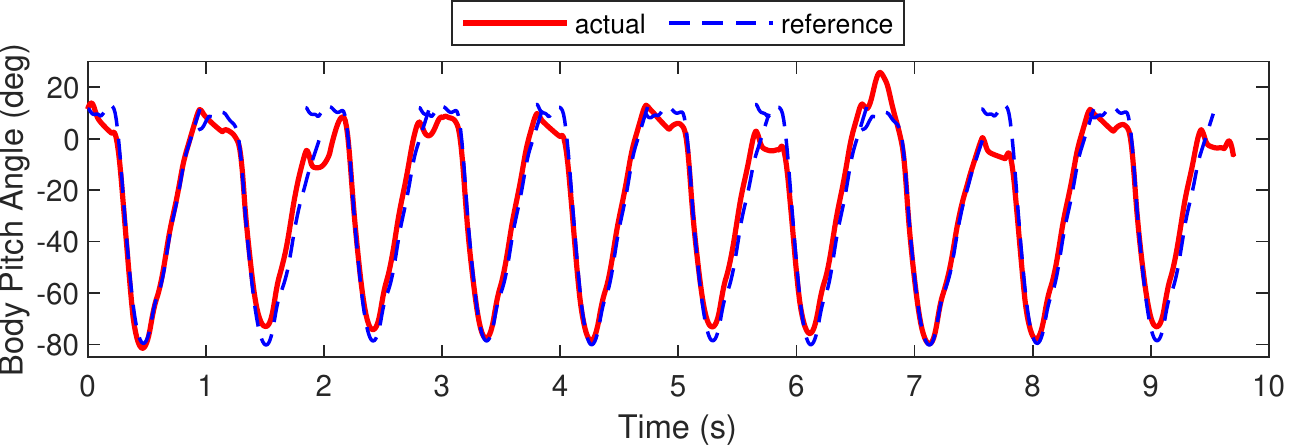}}\\
	\caption{\textbf{Case III- Tracking performance of our controllers with unknown mass.} The robot continuously jumps on the stepping stones while carrying a mass of $2(kg)$, which is unknown to the controllers} 
	\label{fig:Jumpingperformance_unknownmass}
\end{figure}

Note that for Case II \& III, we utilize the same controller parameters as case I to validate the robustness of our framework to unknown disturbance.  Fig.\ref{fig:Jumpingperformance_unknownmass} shows our controller is capable of tracking the references, ensuring continuous jumping on stepping stones with unknown carrying mass.

\subsection{Experimental Verification}
\begin{figure*}[!t]
	\centering
	\subfloat[Baseline experiment]{\centering
		\resizebox{0.33\linewidth}{!}{\includegraphics[trim={0cm 0cm 0cm 0cm},clip]{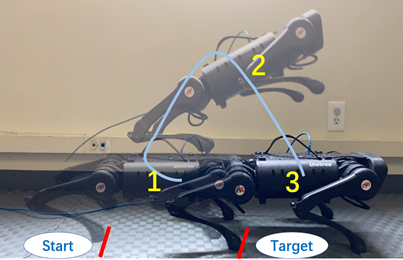}}\label{fig:compare_MPC_experiments_1}
	}
	\subfloat[Joint PD controller]{\centering
		\resizebox{0.33\linewidth}{!}{\includegraphics[trim={0cm 0cm 0cm 0cm},clip]{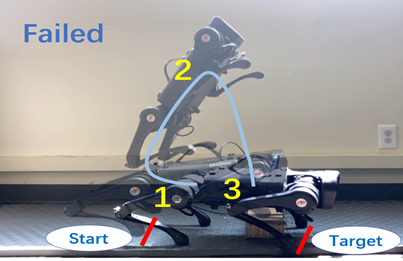}}\label{fig:compare_MPC_experiments_2}
	}
	\subfloat[Our controller]{\centering
		\resizebox{0.33\linewidth}{!}{\includegraphics[trim={0cm 0cm 0cm 0cm},clip]{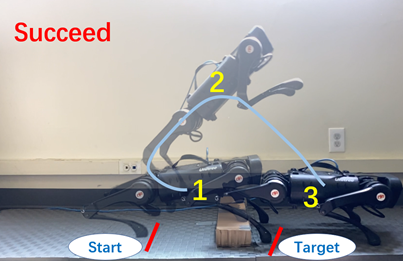}}\label{fig:compare_MPC_experiments_3}
	}\\
	\caption{\textbf{Experiments}: Motion snapshots from jumping forward with (a) baseline experiment: flat ground + joint PD controller , (b) uneven platform + joint PD controller, and (c) uneven platform + our controller. Video available: \url{https://www.youtube.com/watch?v=jBGY1K1UbhM}} 
	\label{fig:compare_MPC_experiments}
\end{figure*}




We demonstrate experiments to verify the robustness of our jumping controller for a single jump. The following experiments aim to show the robustness of our controller to jumping from an uneven platform with unknown height perturbation.

\begin{itemize}
    \item \textit{Baseline experiment} (see Fig.\ref{fig:compare_MPC_experiments_1}): The robot stands up and jumps forward $60cm$ from a predefined initial configuration. 
    \item \textit{Experimental comparisons} (see Fig.\ref{fig:compare_MPC_experiments_2}, \ref{fig:compare_MPC_experiments_3}): We consider a perturbation to the initial configuration (e.g. a box under the front feet), which is unknown to controllers.
\end{itemize}

For the baseline experiment, the reference for jumping is computed from trajectory optimization, then tracked by the joint PD controller in \cite{QuannICRA19},\cite{chuong_jumping_3D_2021}.

For the experimental comparisons, we put a box of $5cm$ in height under the front feet. The height of the box is about $33\%$ of the initial height of the robot. We compare the performance of our proposed controller with the joint PD controller based on actual jumping distances to see if the rear legs traverse over the box and reach the target.

As we can see in Fig. \ref{fig:compare_MPC_experiments_2}, simply utilizing the same joint PD controller as the baseline experiment causes the jumping to fail. This is due to the fact that the joint PD controller works only in the joint space, and there is no control feedback for body orientation and positions. Therefore, it can not compensate for the errors of the jumping trajectory when adding the unknown disturbance. As a result, it accumulates significant errors before taking off, and can only jump a very short distance.
On the other hand, our controller combines MPC and a high frequency PD controller at the joint level, considering real time feedback from body orientation and body position, as well as jumping references from optimization. This combination guarantees high tracking performance while compensating for the unknown disturbance, as illustrated in Fig. \ref{fig:compare_MPC_experiments_3}. 




\section{Conclusion and Future work}
We have presented a framework that combines full-body trajectory optimization and model predictive control to achieve robust and continuous jumping on uneven stepping stones. 
Our proposed framework is validated under various conditions: random distance of stepping stones, and its combination with unknown height perturbations of the platform and unknown mass that the robot is carrying. In addition, we also conduct hardware experiments  to illustrate the robustness of our controller for \edit{a single jump}. Our future work will realize the continuous jumping on stepping stones on the robot hardware.
\label{sec:Conclusion}
\balance
\bibliographystyle{ieeetr}
\bibliography{my_reference}
\end{document}